\def\year{2018}\relax
\documentclass[letterpaper]{article} 
\usepackage{aaai18}  
\usepackage{times}  
\usepackage{helvet}  
\usepackage{courier}  
\usepackage{url}  
\usepackage{graphicx}  
\usepackage{latexsym}
\usepackage{bm}
\usepackage{amsmath}
\usepackage{amssymb}
\usepackage{amsfonts}
\usepackage{multirow}
\usepackage{booktabs}
\usepackage[table]{xcolor}

\title{Deep Semantic Role Labeling with Self-Attention}
\author{
Zhixing Tan\textsuperscript{1},
Mingxuan Wang\textsuperscript{2},
Jun Xie\textsuperscript{2},
Yidong Chen\textsuperscript{1},
Xiaodong Shi\textsuperscript{1}\thanks{\ \ Corresponding author.}\\
\textsuperscript{1}{School of Information Science and Engineering, Xiamen University, Xiamen, China} \\
\textsuperscript{2}{Mobile Internet Group, Tencent Technology Co., Ltd, Beijing, China} \\
\texttt{playinf@stu.xmu.edu.cn}, \\
\{\texttt{xuanswang, stiffxie}\}\texttt{@tencent.com}, \\
\texttt{\{ydchen, mandel\}@xmu.edu.cn} \\
}

\newcolumntype{Y}{>{\centering\arraybackslash}m{0.85cm}}

\begin{document}

\maketitle

\begin{abstract}
Semantic Role Labeling (SRL) is believed to be a crucial step towards natural language understanding and has been widely studied. Recent years, end-to-end SRL with recurrent neural networks (RNN) has gained increasing attention. However, it remains a major challenge for RNNs to handle structural information and long range dependencies. In this paper, we present a simple and effective architecture for SRL which aims to address these problems. Our model is based on self-attention which can directly capture the relationships between two tokens regardless of their distance. Our single model achieves F$_1=83.4$ on the CoNLL-2005 shared task dataset and F$_1=82.7$ on the CoNLL-2012 shared task dataset, which outperforms the previous state-of-the-art results by $1.8$ and $1.0$ F$_1$ score respectively. Besides, our model is computationally efficient, and the parsing speed is 50K tokens per second on a single Titan X GPU.
\end{abstract}

\section{Introduction}
Semantic Role Labeling is a shallow semantic parsing task, whose goal is to determine essentially ``who did what to whom'', ``when'' and ``where''. Semantic roles indicate the basic event properties and relations among relevant entities in the sentence and provide an intermediate level of semantic representation thus benefiting many NLP applications, such as Information Extraction~\cite{Bastianelli-Basili-2013}, Question Answering~\cite{Surdeanu-Aarseth-ACL2003,Moschitti-Harabagiu-FLAIRS2003,Shen-Lapata-CONLL2007}, Machine Translation~\cite{knight1994building,ueffing2007transductive,wu2009semantic} and Multi-document Abstractive Summarization~\cite{genest2011framework}.

Semantic roles are closely related to syntax. Therefore, traditional SRL approaches rely heavily on the syntactic structure of a sentence, which brings intrinsic complexity and restrains these systems to be domain specific. Recently, end-to-end models for SRL without syntactic inputs achieved promising results on this task~\cite{zhou2015end,marcheggiani2017simple,he2017deep}. As the pioneering work, Zhou and Xu~\shortcite{zhou2015end} introduced a stacked long short-term memory network (LSTM) and achieved the state-of-the-art results. He et al.,~\shortcite{he2017deep} reported further improvements by using deep highway bidirectional LSTMs with constrained decoding. These successes involving end-to-end models reveal the potential ability of LSTMs for handling the underlying syntactic structure of the sentences.

Despite recent successes, these RNN-based models have limitations. RNNs treat each sentence as a sequence of words and recursively compose each word with its previous hidden state. The recurrent connections make RNNs applicable for sequential prediction tasks with arbitrary length, however, there still remain several challenges in practice. The first one is related to memory compression problem~\cite{cheng2016long}. As the entire history is encoded into a single fixed-size vector, the model requires larger memory capacity to store information for longer sentences. The unbalanced way of dealing with sequential information leads the network performing poorly on long sentences while wasting memory on shorter ones. The second one is concerned with the inherent structure of sentences. RNNs lack a way to tackle the tree-structure of the inputs. The sequential way to process the inputs remains the network depth-in-time, and the number of nonlinearities depends on the time-steps.

To address these problems above, we present a deep attentional neural network (\textsc{DeepAtt}) for the task of SRL\footnote{Our source code is available at \url{https://github.com/XMUNLP/Tagger}}. Our models rely on the self-attention mechanism which directly draws the global dependencies of the inputs. In contrast to RNNs, a major advantage of self-attention is that it conducts direct connections between two arbitrary tokens in a sentence. Therefore, distant elements can interact with each other by shorter paths $\left(O(1) \ \textrm{v.s.}\  O(n)\right)$, which allows unimpeded information flow through the network. Self-attention also provides a more flexible way to select, represent and synthesize the information of the inputs and is complementary to RNN based models. Along with self-attention, \textsc{DeepAtt} comes with three variants which uses recurrent (RNN), convolutional (CNN) and feed-forward (FFN) neural network to further enhance the representations.

Although \textsc{DeepAtt} is fairly simple, it gives remarkable empirical results. Our single model outperforms the previous state-of-the-art systems on the CoNLL-2005 shared task dataset and the CoNLL-2012 shared task dataset by $1.8$ and $1.0$ F$_1$ score respectively. It is also worth mentioning that on the out-of-domain dataset, we achieve an improvement upon the previous end-to-end approach~\cite{he2017deep} by $2.0$ F$_1$ score. The feed-forward variant of \textsc{DeepAtt} allows significantly more parallelization, and the parsing speed is 50K tokens per second on a single Titan X GPU.

\section{Semantic Role Labeling}
Given a sentence, the goal of SRL is to identify and classify the arguments of each target verb into semantic roles. For example, for the sentence \textit{``Marry borrowed a book from John last week.''} and the target verb \textit{borrowed}, SRL yields the following outputs:

\begin{verse}
$[_{\textrm{ARG0}}$  Marry $] $
$[_{\textrm{V}}$  borrowed $] $
$[_{\textrm{ARG1}}$ a book $]  $ \\
$[_{\textrm{ARG2}}$ from John $] $
$[_{\textrm{AM-TMP}}$  last week $]. $
\end{verse}

Here ARG0 represents the \textit{borrower}, ARG1 represents the \textit{thing borrowed}, ARG2 represents the \textit{entity borrowed from}, AM-TMP is an adjunct indicating the timing of the action and V represents the verb.

Generally, semantic role labeling consists of two steps: identifying and classifying arguments. The former step involves assigning either a semantic argument or non-argument for a given predicate, while the latter includes labeling a specific semantic role for the identified argument. It is also common to prune obvious non-candidates before the first step and to apply post-processing procedure to fix inconsistent predictions after the second step. Finally, a dynamic programming algorithm is often applied to find the global optimum solution for this typical sequence labeling problem at the inference stage.

\begin{figure}[!ht]
\begin{center}
  \includegraphics[width=0.45\textwidth]{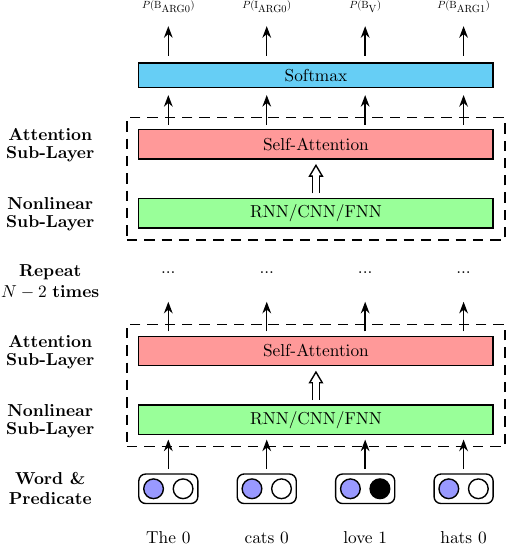}
\end{center}
\caption{An illustration of our deep attentional neural network. Original utterances and corresponding predicate masks are taken as the only inputs for our deep model. For example, \textit{love} is the predicate and marked as $1$, while other words are marked as $0$.}
\label{f:overall}
\end{figure}

In this paper, we treat SRL as a BIO tagging problem. Our approach is extremely simple. As illustrated in Figure \ref{f:overall}, the original utterances and the corresponding predicate masks are first projected into real-value vectors, namely embeddings, which are fed to the next layer. After that, we design a deep attentional neural network which takes the embeddings as the inputs to capture the nested structures of the sentence and the latent dependency relationships among the labels. On the inference stage, only the topmost outputs of attention sub-layer are taken to a logistic regression layer to make the final decision~\footnote{In case of BIO violations, we simply treat the argument of the B tags as the argument of the whole span.}.

\section{Deep Attentional Neural Network for SRL}
In this section, we will describe \textsc{DeepAtt} in detail. The main component of our deep network consists of $N$ identical layers. Each layer contains a nonlinear sub-layer followed by an attentional sub-layer. The topmost layer is the softmax classification layer.

\subsection{Self-Attention}
Self-attention or intra-attention, is a special case of attention mechanism that only requires a single sequence to compute its representation. Self-attention has been successfully applied to many tasks, including reading comprehension, abstractive summarization, textual entailment, learning task-independent sentence representations, machine translation and language understanding~\cite{cheng2016long,parikh2016decomposable,lin2017structured,paulus2017deep,vaswani2017attention,shen2017disan}.

\begin{figure*}[!ht]
\centering
\includegraphics[width=0.7\textwidth]{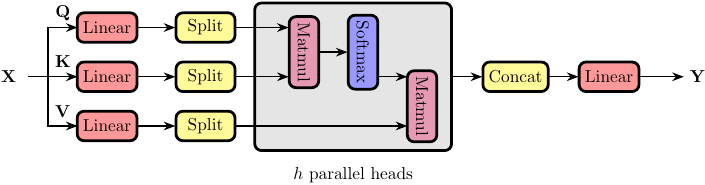}
\caption{The computation graph of multi-head self-attention mechanism. All heads can be computed in parallel using highly optimized matrix multiplication codes.}
\label{fig:mhatt}
\end{figure*}

In this paper, we adopt the multi-head attention formulation by Vaswani et al.~\shortcite{vaswani2017attention}. Figure \ref{fig:mhatt} depicts the computation graph of multi-head attention mechanism. The center of the graph is the scaled dot-product attention, which is a variant of dot-product (multiplicative) attention~\cite{luong2015effective}. Compared with the standard additive attention mechanism~\cite{bahdanau2014neural} which is implemented using a one layer feed-forward neural network, the dot-product attention utilizes matrix production which allows faster computation. Given a matrix of $n$ query vectors $\mathbf{Q} \in \mathbb{R}^{n \times d}$, keys $\mathbf{K} \in \mathbb{R}^{n \times d}$ and values $\mathbf{V} \in \mathbb{R}^{n \times d}$, the scaled dot-product attention computes the attention scores based on the following equation:
\begin{equation}
\textrm{Attention}(\mathbf{Q}, \mathbf{K}, \mathbf{V}) = \textrm{softmax}(\frac{\mathbf{Q}\mathbf{K}^T}{\sqrt{d}})\mathbf{V}
\end{equation}
where $d$ is the number of hidden units of our network.

The multi-head attention mechanism first maps the matrix of input vectors $\mathbf{X} \in \mathbb{R}^{t \times d}$ to queries, keys and values matrices by using different linear projections. Then $h$ parallel heads are employed to focus on different part of channels of the value vectors. Formally, for the $i$-th head, we denote the learned linear maps by $\mathbf{W}_i^Q \in \mathbb{R}^{n \times d/h}$, $\mathbf{W}_i^K \in \mathbb{R}^{n \times d/h}$ and $\mathbf{W}_i^V \in \mathbb{R}^{n \times d/h}$, which correspond to queries, keys and values respectively. Then the scaled dot-product attention is used to compute the relevance between queries and keys, and to output mixed representations. The mathematical formulation is shown below:
\begin{equation}
\mathbf{M}_i = \textrm{Attention}(\mathbf{Q}\mathbf{W}_i^Q, \mathbf{K}\mathbf{W}_i^K, \mathbf{V}\mathbf{W}_i^V)
\end{equation}

Finally, all the vectors produced by parallel heads are concatenated together to form a single vector. Again, a linear map is used to mix different channels from different heads:
\begin{align}
\mathbf{M} &= \textrm{Concat}(\mathbf{M}_1, \ldots, \mathbf{M}_h) \\
\mathbf{Y} &= \mathbf{M}\mathbf{W}
\end{align}
where $\mathbf{M} \in \mathbb{R}^{n \times d}$ and $\mathbf{W} \in \mathbb{R}^{d \times d}$.

The self-attention mechanism has many appealing aspects compared with RNNs or CNNs. Firstly, the distance between any input and output positions is 1, whereas in RNNs it can be $n$. Unlike CNNs, self-attention is not limited to fixed window sizes. Secondly, the attention mechanism uses weighted sum to produce output vectors. As a result, the gradient propagations are much easier than RNNs or CNNs. Finally, the dot-product attention is highly parallel. In contrast, RNNs are hard to parallelize owing to its recursive computation.

\subsection{Nonlinear Sub-Layers}
The successes of neural networks root in its highly flexible nonlinear transformations. Since attention mechanism uses weighted sum to generate output vectors, its representational power is limited. To further increase the expressive power of our attentional network, we employ a nonlinear sub-layer to transform the inputs from the bottom layers. In this paper, we explore three kinds of nonlinear sub-layers, namely recurrent, convolutional and feed-forward sub-layers.

\subsubsection{Recurrent Sub-Layer}
We use bidirectional LSTMs to build our recurrent sub-layer. Given a sequence of input vectors $\{\mathbf{x}_t\}$, two LSTMs process the inputs in opposite directions. To maintain the same dimension between inputs and outputs, we use the sum operation to combine two representations:
\begin{align}
  \overrightarrow{\mathbf{h}}_t &= \textrm{LSTM}(\mathbf{x}_t, \overrightarrow{\mathbf{h}}_{t-1})\\
  \overleftarrow{\mathbf{h}}_t &= \textrm{LSTM}(\mathbf{x}_t, \overleftarrow{\mathbf{h}}_{t+1})\\
  \mathbf{y}_t &= \overrightarrow{\mathbf{h}}_t + \overleftarrow{\mathbf{h}}_t
\end{align}

\subsubsection{Convolutional Sub-Layer}
For convolutional sub-layer, we use the Gated Linear Unit (GLU) proposed by Dauphin et al.~\shortcite{dauphin2016language}. Compared with the standard convolutional neural network, GLU is much easier to learn and achieves impressive results on both language modeling and machine translation task~\cite{dauphin2016language,gehring2017convolutional}. Given two filters $\mathbf{W} \in \mathbb{R}^{k \times d \times d}$ and $\mathbf{V} \in \mathbb{R}^{k \times d \times d}$, the output activations of GLU are computed as follows:
\begin{equation}
\textrm{GLU}(\mathbf{X}) = (\mathbf{X}*\mathbf{W}) \odot \sigma(\mathbf{X}*\mathbf{V})
\end{equation}
The filter width $k$ is set to 3 in all our experiments.

\subsubsection{Feed-forward Sub-Layer}
The feed-forward sub-layer is quite simple. It consists of two linear layers with hidden ReLU nonlinearity~\cite{nair2010rectified} in the middle. Formally, we have the following equation:
\begin{equation}
\textrm{FFN}(\mathbf{X}) = \textrm{ReLU}(\mathbf{X}\mathbf{W}_1)\mathbf{W}_2
\end{equation}
where $\mathbf{W}_1 \in \mathbb{R}^{d \times h_f}$ and $\mathbf{W}_2 \in \mathbb{R}^{h_f \times d}$ are trainable matrices. Unless otherwise noted, we set $h_f = 800$ in all our experiments.

\subsection{Deep Topology}
Previous works pointed out that deep topology is essential to achieve good performance~\cite{zhou2015end,he2017deep}. In this work, we use the residual connections proposed by He et al.~\shortcite{he2016deep} to ease the training of our deep attentional neural network. Specifically, the output $\mathbf{Y}$ of each sub-layer is computed by the following equation:
\begin{equation}
\mathbf{Y} = \mathbf{X} + \textrm{Sub-Layer}(\mathbf{X})
\end{equation}
We then apply layer normalization~\cite{ba2016layer} after the residual connection to stabilize the activations of deep neural network.

\subsection{Position Encoding}
The attention mechanism itself cannot distinguish between different positions. So it is crucial to encode positions of each input words. There are various ways to encode positions, and the simplest one is to use an additional position embedding. In this work, we try the timing signal approach proposed by Vaswani et al.~\shortcite{vaswani2017attention}, which is formulated as follows:
\begin{align}
\textrm{timing}(t, 2i) = \sin(t / 10000^{2i / d}) \\
\textrm{timing}(t, 2i + 1) = \cos(t / 10000^{2i / d})
\end{align}
The timing signals are simply added to the input embeddings. Unlike the position embedding approach, this approach does not introduce additional parameters.

\subsection{Pipeline}
The first step of using neural networks to process symbolic data is to represent them by distributed vectors, also called embeddings~\cite{Bengio-Janvin-JMLR2003}. We take the very original utterances and the corresponding predicate masks $\mathbf{m}$ as the input features. $m_t$ is set to $1$ if the corresponding word is a predicate, or $0$ if not.

Formally, in SRL task, we have a word vocabulary $\mathcal{V}$ and mask vocabulary $\mathcal{C}=\{0,1\}$. Given a word sequence $\{x_1, x_2, \ldots, x_T\}$ and a mask sequence $\{m_1, m_2, ..., m_T\}$, each word $x_t \in \mathcal{V}$ and its corresponding predicate mask $m_t \in \mathcal{C}$ are projected into real-valued vectors $\mathbf{e}(x_t)$ and $\mathbf{e}(m_t)$ through the corresponding lookup table layer, respectively. The two embeddings are then concatenated together as the output feature maps of the lookup table layers. Formally speaking, we have $\mathbf{x}_t = \left[\mathbf{e}(w_t),\mathbf{e}(m_t)\right]$.

We then build our deep attentional neural network to learn the sequential and structural information of a given sentence based on the feature maps from the lookup table layer. Finally, we take the outputs of the topmost attention sub-layer as inputs to make the final predictions.

Since there are dependencies between semantic labels, most previous neural network models introduced a transition model for measuring the probability of jumping between the labels. Different from these works, we perform SRL as a typical classification problem. Latent dependency information is embedded in the topmost attention sub-layer learned by our deep models. This approach is simpler and easier to implement compared to previous works.

Formally, given an input sequence $\bm{x} = \{x_1, x_2,\ldots,x_n\}$, the log-likelihood of the corresponding correct label sequence $\bm{y} = \{y_1, y_2, \ldots, y_n\}$ is
\begin{equation}
\log p(\bm{y} | \bm{x}; \bm{\theta}) = \sum_{t=1}^{n} \log p(y_t|\bm{x}; \bm{\theta}).
\end{equation}
Our model predict the corresponding label $y_t$ based on the representation $\mathbf{h}_t$ produced by the topmost attention sub-layer of \textsc{DeepAtt}:
\begin{eqnarray}
    p(y_t|\bm{x}; \bm{\theta}) &=& p(y_t|\mathbf{h}_t; \bm{\theta}) \\
    &=& \textrm{softmax}(\mathbf{W}_o\mathbf{h}_t)^T\delta_{y_t},
\end{eqnarray}
Where $W_o$ is the softmax matrix and $\delta_{y_t}$ is Kronecker delta with a dimension for each output symbol, so $\textrm{softmax}(\mathbf{W}_o\mathbf{h}_t)^T\delta_{y_t}$ is exactly the $y_t$'th element of the distribution defined by the softmax. Our training objective is to maximize the log probabilities of the correct output labels given the input sequence over the entire training set.

\section{Experiments}
We report our empirical studies of \textsc{DeepAtt} on the two commonly used datasets from the CoNLL-2005 shared task and the CoNLL-2012 shared task.

\subsection{Datasets}
The CoNLL-2005 dataset takes section 2-21 of the Wall Street Journal (WSJ) corpus as training set, and section 24 as development set. The test set consists of section 23 of the WSJ corpus as well as 3 sections from the Brown corpus~\cite{Carreras-CoNLL2005}. The CoNLL-2012 dataset is extracted from the OntoNotes v5.0 corpus. The description and separation of training, development and test set can be found in Pardhan et al.~\shortcite{Pradhan-CoNLL2013}.

\subsection{Model Setup}
\paragraph{Initialization} We initialize the weights of all sub-layers as random orthogonal matrices. For other parameters, we initialize them by sampling each element from a Gaussian distribution with mean $0$ and variance $\frac{1}{\sqrt{d}}$. The embedding layer can be initialized randomly or using pre-trained word embeddings. We will discuss the impact of pre-training in the analysis subsection.\footnote{To be strictly comparable to previous work, we use the same vocabularies and pre-trained embeddings as He et al.\shortcite{he2017deep}.}

\paragraph{Settings and Regularization} The settings of our models are described as follows. The dimension of word embeddings and predicate mask embeddings is set to 100 and the number of hidden layers is set to 10. We set the number of hidden units $d$ to $200$. The number of heads $h$ is set to 8. We apply dropout~\cite{srivastava2014dropout} to prevent the networks from over-fitting. Dropout layers are added before residual connections with a keep probability of 0.8. Dropout is also applied before the attention softmax layer and the feed-froward ReLU hidden layer, and the keep probabilities are set to 0.9. We also employ label smoothing technique~\cite{szegedy2016rethinking} with a smoothing value of 0.1 during training.

\paragraph{Learning}
Parameter optimization is performed using stochastic gradient descent. We adopt Adadelta \cite{zeiler2012adadelta} ($ \epsilon= 10^{−6}$ and $\rho = 0.95$) as the optimizer. To avoid exploding gradients problem, we clip the norm of gradients with a predefined threshold $1.0$~\cite{pascanu2013construct}. Each SGD contains a mini-batch of approximately 4096 tokens for the CoNLL-2005 dataset and 8192 tokens for the CoNLL-2012 dataset. The learning rate is initialized to 1.0. After training 400k steps, we halve the learning rate every 100K steps. We train all models for 600K steps. For \textsc{DeepAtt} with FFN sub-layers, the whole training stage takes about two days to finish on a single Titan X GPU, which is 2.5 times faster than the previous approach~\cite{he2017deep}.

\subsection{Results}
\begin{table*}[ht]
\centering
\resizebox{\linewidth}{!}{%
\begin{tabular}{cccccccccccccc}
\hline
\multirow{2}{*}{Model} & \multicolumn{4}{c}{Development} & \multicolumn{4}{c}{WSJ Test} & \multicolumn{4}{c}{Brown Test} & {Combined} \\ \cmidrule(lr){2-5} \cmidrule(lr){6-9} \cmidrule(lr){10-13} \cmidrule(lr){14-14}
                       & P      & R      & F1      & Comp.  & P     & R     & F1     & Comp.  & P      & R      & F1     & Comp.  & F1        \\ \hline
He et al. (Ensemble)~\shortcite{he2017deep} & 83.1   & 82.4  & 82.7  & 64.1 & 85.0 & 84.3 & 84.6 & 66.5 & \textbf{74.9}  & 72.4  & 73.6 & 46.5 & 83.2         \\
He et al. (Single)~\shortcite{he2017deep} & 81.6  & 81.6  & 81.6  & 62.3 & 83.1 & 83.0 & 83.1 & 64.3 & 72.8  & 71.4  & 72.1 & 44.8 & 81.6         \\
Zhou and Xu~\shortcite{zhou2015end} & 79.7  & 79.4  & 79.6  & - & 82.9 & 82.8 & 82.8 & - & 70.7  & 68.2  & 69.4 & - & 81.1         \\
FitzGerald et al. (Struct., Ensemble)~\shortcite{fitzgerald2015semantic}      & 81.2  & 76.7  & 78.9  & 55.1 & 82.5 & 78.2 & 80.3 & 57.3 & 74.5  & 70.0  & 72.2 & 41.3 & -         \\
T\"{a}ckstr\"{o}m et al. (Struct.)~\shortcite{Tackstrom-Das-TACL2015}      & 81.2  & 76.2  & 78.6  & 54.4 & 82.3 & 77.6 & 79.9 & 56.0 & 74.3  & 68.6  & 71.3 & 39.8 & -         \\
Toutanova et al. (Ensemble)~\shortcite{Toutanova-Manning-CL2008}      & -  & -  & 78.6  & 58.7 & 81.9 & 78.8 & 80.3 & 60.1 & -  & -  & 68.8 & 40.8 & -         \\
Punyakanok et al. (Ensemble)~\shortcite{Punyakanok-Yih-2008}      & 80.1  & 74.8  & 77.4  & 50.7 & 82.3 & 76.8 & 79.4 & 53.8 & 73.4  & 62.9  & 67.8 & 32.3 & 77.9         \\\hline
\textsc{DeepAtt} (RNN)     & 81.2  & 82.3  & 81.8  & 62.4 & 83.5 & 84.0 & 83.7 & 65.2 & 72.5  & 73.4  & 72.9 & 44.7 & 82.3 \\
\textsc{DeepAtt} (CNN)     & 82.1  & 82.8  & 82.4  & 63.6 & 83.6 & 83.9 & 83.8 & 65.4 & 72.8  & 72.7  & 72.7 & 45.9 & 82.3 \\
\textsc{DeepAtt} (FFN)     & 82.6  & 83.6  & 83.1  & 65.2 & 84.5 & 85.2 & 84.8 & 66.4 & 73.5  & 74.6  & 74.1 & 48.4 & 83.4 \\
\textsc{DeepAtt} (FFN, Ensemble)       & \textbf{84.3} & \textbf{84.9} & \textbf{84.6} &\textbf{67.3} & \textbf{85.9} & \textbf{86.3} & \textbf{86.1} & \textbf{69.0} & 74.6 & \textbf{75.0} & \textbf{74.8} & \textbf{48.6} & \textbf{84.6} \\ \hline
\end{tabular}
}
\caption{\label{tab:Eval-05} Comparison with previous methods on the CoNLL-2005 dataset. We report the results
in terms of precision (P), recall (R), F$_1$ and percentage of completely correct predicates (Comp.). Our single and ensemble model lead to substantial improvements over the previous state-of-the-art results.}
\end{table*}

\begin{table*}[ht]
\centering
\resizebox{0.8\linewidth}{!}{%
\begin{tabular}{ccccccccc}
\hline
\multirow{2}{*}{Model} & \multicolumn{4}{c}{Development} & \multicolumn{4}{c}{Test} \\ \cmidrule(lr){2-5} \cmidrule(lr){6-9}
                       & P      & R      & F1      & Comp.  & P     & R     & F1     & Comp.  \\ \hline
He et al. (Ensemble)~\shortcite{he2017deep} & 83.5  & 83.2  & 83.4  & 67.5   & \textbf{83.5} & 83.3 & 83.4 & 68.5 \\
He et al. (Single)~\shortcite{he2017deep} & 81.7  & 81.4  & 81.5  & 64.6 & 81.8 & 81.6 & 81.7 & 66.0 \\
Zhou and Xu~\shortcite{zhou2015end} & -  & -  & 81.1  & - & - & - & 81.3 & - \\
FitzGerald et al. (Struct., Ensemble)~\shortcite{fitzgerald2015semantic}  & 81.0  & 78.5  & 79.7  & 60.9 & 81.2 & 79.0 & 80.1 & 62.6 \\
T\"{a}ckstr\"{o}m et al. (Struct., Ensemble)~\shortcite{Tackstrom-Das-TACL2015}  & 80.5  & 77.8  & 79.1  & 60.1 & 80.6 & 78.2 & 79.4 & 61.8 \\
Pradhan et al.(Revised)~\shortcite{Pradhan-CoNLL2013}  & -  & -  & -  & - & 78.5 & 76.6 & 77.5 & 55.8 \\ \hline
\textsc{DeepAtt} (RNN) & 81.0 & 82.3 & 81.6 & 64.6  & 80.9 & 82.2 & 81.5 & 65.7 \\
\textsc{DeepAtt} (CNN) & 80.1  & 82.5  & 81.3  & 65.0 & 79.8 & 82.6 & 81.2 & 66.1 \\
\textsc{DeepAtt} (FFN) & 82.2  & 83.6  & 82.9  & 66.7 & 81.9 & 83.6 & 82.7 & 67.5 \\
\textsc{DeepAtt} (FFN, Ensemble) & \textbf{83.6} & \textbf{84.7}  & \textbf{84.1} & \textbf{68.7} & 83.3 & \textbf{84.5} & \textbf{83.9} & \textbf{69.3} \\ \hline
\end{tabular}}
\caption{\label{tab:Eval-12} Experimental results on the CoNLL-2012 dataset. The metrics are the same as above. Again, our model achieves the state-of-the-art performance.}
\end{table*}

In Table \ref{tab:Eval-05} and \ref{tab:Eval-12}, we give the comparisons of \textsc{DeepAtt} with previous approaches.  On the CoNLL-2005 dataset, the single model of \textsc{DeepAtt} with RNN, CNN and FFN nonlinear sub-layers achieves an F$_1$ score of $82.3$, $82.3$ and $83.4$ respectively. The FFN variant outperforms previous best performance by 1.8 F$_1$ score. Remarkably, we get 74.1 F$_1$ score on the out-of-domain dataset, which outperforms the previous state-of-the-art system by $2.0$ F$_1$ score. On the CoNLL-2012 dataset, the single model of FFN variant also outperforms the previous state-of-the-art by 1.0 F$_1$ score. When ensembling 5 models with FFN nonlinear sub-layers, our approach achieves an F$_1$ score of 84.6 and 83.9 on the two datasets respectively, which has an absolute improvement of 1.4 and 0.5 over the previous state-of-the-art. These results are consistent with our intuition that the self-attention layers is helpful to capture structural information and long distance dependencies.

\subsection{Analysis}
In this subsection, we discuss the main factors that influence our results. We analyze the experimental results on the development set of CoNLL-2005 dataset.

\begin{table}[htb]
\centering
\resizebox{\linewidth}{!}{%
\begin{tabular}{llllcc|c}
\hline
\multicolumn{7}{c}{CoNLL-2005 Dataset} \\\hline\hline
\# & Nonlinearity & PE   &Embedding & Width & Depth & F$_1$ \\\hline
1 & FFN & Timing & GloVe & 200 & 10 & 83.1 \\\hline
2 & FFN & Timing & GloVe & 200 &4&  79.9 \\
3 & FFN & Timing & GloVe & 200 &6&  82.0 \\
4 & FFN & Timing & GloVe & 200 &8&  82.8 \\
5 & FFN & Timing & GloVe & 200 &12&  83.0 \\\hline
6 & FFN & Timing & GloVe & 400 & 10& 83.2 \\
7 & FFN & Timing & GloVe & 600 & 10 & 83.4 \\\hline
8 & FFN &  Timing & Random & 200 &10&  79.6 \\\hline
9 & FFN & None   & GloVe  & 200 &10& 20.0 \\
10 & FFN & Embedding & GloVe & 200 & 10& 79.4 \\\hline
11 & None & Timing & GloVe & 200 & 10 & 79.9 \\\hline
\end{tabular}}
\caption{\label{tab:EXP-Group} Detailed results on the CoNLL-2005 development set. \textsc{PE} denotes the way to encoding word positions. GloVe refers to the GloVe embedding pre-trained on 6B tokens.}
\end{table}

\paragraph{Model Depth}
Previous works~\cite{zhou2015end,he2017deep} show that model depth is the key to the success of end-to-end SRL approach. Our observations also coincide with previous works. Rows 1-5 of Table \ref{tab:EXP-Group} show the effects of different number of layers. For \textsc{DeepAtt} with 4 layers, our model only achieves 79.9 F$_1$ score. Increasing depth consistently improves the performance on the development set, and our best model consists of 10 layers. For \textsc{DeepAtt} with 12 layers, we observe a slightly performance drop of 0.1 F$_1$.

\paragraph{Model Width} We also conduct experiments with different model widths. We increase the number of hidden units from $200$ to $400$ and $400$ to $600$ as listed in rows 1, 6 and 7 of Table \ref{tab:EXP-Group}, and the corresponding hidden size $h_f$ of FFN sub-layers is increased to 1600 and 2400 respectively. Increasing model widths improves the F$_1$ slightly, and the model with 600 hidden units achieves an F$_1$ of 83.4. However, the training and parsing speed are slower as a result of larger parameter counts.

\paragraph{Word Embedding}
Previous works found that the performance can be improved by pre-training the word embeddings on large unlabeled data~\cite{Collobert-Ronan-JMLR2011,zhou2015end}. We use the GloVe~\cite{pennington2014glove} embeddings pre-trained on Wikipedia and Gigaword. The embeddings are used to initialize our networks, but are not fixed during training. Rows 1 and 8 of Table \ref{tab:EXP-Group} show the effects of additional pre-trained embeddings. When using pre-trained GloVe embeddings, the F$_1$ score increases from 79.6 to 83.1.

\paragraph{Position Encoding}
From rows 1, 9 and 10 of Table \ref{tab:EXP-Group} we can see that the position encoding plays an important role in the success of \textsc{DeepAtt}. Without position encoding, the \textsc{DeepAtt} with FFN sub-layers only achieves $20.0$ F$_1$ score on the CoNLL-2005 development set. When using position embedding approach, the F$_1$ score boosts to $79.4$. The timing approach is surprisingly effective, which outperforms the position embedding approach by $3.7$ F$_1$ score.

\paragraph{Nonlinear Sub-Layers} \textsc{DeepAtt} requires nonlinear sub-layers to enhance its expressive power. Row 11 of Table \ref{tab:EXP-Group} shows the performance of \textsc{DeepAtt} without nonlinear sub-layers. We can see that the performance of 10 layered \textsc{DeepAtt} without nonlinear sub-layers only matches the 4 layered \textsc{DeepAtt} with FFN sub-layers, which indicates that the nonlinear sub-layers are the essential components of our attentional networks.

\begin{table}[!t]
\centering
\begin{tabular}{c|cc}
Decoding & F$_1$ & Speed \\\hline
Argmax Decoding & \textbf{83.1} & \textbf{50K} \\
Constrained Decoding & 83.0 & 17K
\end{tabular}
\caption{Comparison between argmax decoding and constrained decoding on top of our model.}
\label{EXP:Decoding}
\end{table}

\paragraph{Constrained Decoding} Table \ref{EXP:Decoding} show the effects of constrained decoding~\cite{he2017deep} on top of \textsc{DeepAtt} with FFN sub-layers. We observe a slightly performance drop when using constrained decoding. Moreover, adding constrained decoding slow down the decoding speed significantly. For \textsc{DeepAtt}, it is powerful enough to capture the relationships among labels.

\begin{table}[!ht]
\centering
\resizebox{0.45\textwidth}{!}{%
\begin{tabular}{lcccccc}\cline{1-7}
\multirow{2}{*}{Label} & \multicolumn{3}{c}{He et al.~\shortcite{he2017deep}} & \multicolumn{3}{c}{\textsc{DeepAtt}}\\
\cmidrule(lr){2-4} \cmidrule(lr){5-7}
 & Precision & Recall & F$_1$ & Precision & Recall & F$_1$\\\hline
A0                 &  89.6 &  90.7 &  90.2 &  91.0 &  91.9 &  91.4\\
A1                 &  84.2 &  84.6 &  84.4 &  86.2 &  86.9 &  86.6\\
A2                 &  73.6 &  72.7 &  73.2 &  75.0 &  77.0 &  76.0\\
A3                &  76.6 &  63.2 &  69.2 &  77.1 &  73.7 &  75.3\\
AM-ADV             &  64.8 &  60.6 &  62.6 &  69.5 &  64.5 &  66.9\\
AM-DIR            &  48.3 &  38.9 &  43.1 &  54.8 &  47.2 &  50.8\\
AM-LOC            &  59.8 &  58.3 &  59.0 &  62.5 &  61.9 &  62.2\\
AM-MNR             &  70.4 &  57.0 &  63.0 &  70.4 &  62.0 &  65.9\\
AM-PNC             &  65.8 &  64.2 &  65.0 &  65.3 &  60.5 &  62.8\\
AM-TMP             &  80.5 &  85.7 &  83.0 &  81.7 &  87.9 &  84.7\\
\hline
Overall            &  83.1 &  82.4 &  82.7 &  84.3 &  84.9 &  84.6\\
\hline
\end{tabular}
}
\caption{Detailed scores on the development set of CoNLL-2005 dataset. We also list the previous state-of-the-art model~\cite{he2017deep} for comparison.}
\label{EXP:detailed}
\end{table}

\begin{table}[!t]
\centering
\begin{tabular}{c|cc}
Model & Constituents & Semantic Roles \\\hline
He et al.~\shortcite{he2017deep} & 91.87 & 87.10 \\
\textsc{DeepAtt} & \textbf{91.92} & \textbf{88.88} \\
\end{tabular}
\caption{Comparison with the previous work on identifying and classifying semantic roles. We list the percentage of correctly identified spans as well as the percentage of correctly classified semantic roles given the gold spans.}
\label{EXP:percentage}
\end{table}

\paragraph{Detailed Scores} We list the detailed performance on frequent labels in Table \ref{EXP:detailed}. The results of the previous state-of-the-art~\cite{he2017deep} are also shown for comparison. Compared with He et al.~\shortcite{he2017deep}, our model shows improvement on all labels except AM-PNC, where He's model performs better. Table \ref{EXP:percentage} shows the results of identifying and classifying semantic roles. Our model improves the previous state-of-the-art on both identifying correct spans as well as correctly classifying them into semantic roles. However, the majority of improvements come from classifying semantic roles. This indicates that finding the right constituents remains a bottleneck of our model.

\begin{table}[!ht]
\centering
\resizebox{0.45\textwidth}{!}{%
\begin{tabular}{c *{12}{Y}}
 \multicolumn{1}{c}{pred./gold} & \multicolumn{1}{c}{A0} & \multicolumn{1}{c}{A1}
 & \multicolumn{1}{c}{A2} & \multicolumn{1}{c}{A3} & \multicolumn{1}{c}{ADV}
 & \multicolumn{1}{c}{DIR} & \multicolumn{1}{c}{LOC} & \multicolumn{1}{c}{MNR}
 & \multicolumn{1}{c}{PNC} & \multicolumn{1}{c}{TMP} \\
 A0 &  - \cellcolor[gray]{1.00} & 48 \cellcolor[gray]{0.51} & 12 \cellcolor[gray]{0.88} &  7 \cellcolor[gray]{0.93} &  0 \cellcolor[gray]{1.00} &  0 \cellcolor[gray]{1.00} &  0 \cellcolor[gray]{1.00} &  0 \cellcolor[gray]{1.00} &  7 \cellcolor[gray]{0.92} &  0 \cellcolor[gray]{1.00} \\
 A1 & 76 \cellcolor[gray]{0.23} &  - \cellcolor[gray]{1.00} & 35 \cellcolor[gray]{0.65} &  0 \cellcolor[gray]{1.00} &  7 \cellcolor[gray]{0.92} & 16 \cellcolor[gray]{0.83} & 19 \cellcolor[gray]{0.81} &  2 \cellcolor[gray]{0.98} & 30 \cellcolor[gray]{0.69} &  0 \cellcolor[gray]{1.00} \\
 A2 & 10 \cellcolor[gray]{0.90} & 37 \cellcolor[gray]{0.62} &  - \cellcolor[gray]{1.00} & 42 \cellcolor[gray]{0.57} & 15 \cellcolor[gray]{0.85} & 33 \cellcolor[gray]{0.67} & 28 \cellcolor[gray]{0.71} & 35 \cellcolor[gray]{0.64} & 38 \cellcolor[gray]{0.62} &  0 \cellcolor[gray]{1.00} \\
 A3 &  0 \cellcolor[gray]{1.00} &  0 \cellcolor[gray]{1.00} &  5 \cellcolor[gray]{0.95} &  - \cellcolor[gray]{1.00} &  0 \cellcolor[gray]{1.00} &  0 \cellcolor[gray]{1.00} &  0 \cellcolor[gray]{1.00} &  2 \cellcolor[gray]{0.98} &  7 \cellcolor[gray]{0.92} &  0 \cellcolor[gray]{1.00} \\
ADV &  0 \cellcolor[gray]{1.00} &  0 \cellcolor[gray]{1.00} &  0 \cellcolor[gray]{1.00} &  0 \cellcolor[gray]{1.00} &  - \cellcolor[gray]{1.00} &  0 \cellcolor[gray]{1.00} &  9 \cellcolor[gray]{0.90} & 26 \cellcolor[gray]{0.73} & 15 \cellcolor[gray]{0.85} & 58 \cellcolor[gray]{0.42} \\
DIR &  0 \cellcolor[gray]{1.00} &  0 \cellcolor[gray]{1.00} &  2 \cellcolor[gray]{0.97} &  0 \cellcolor[gray]{1.00} &  0 \cellcolor[gray]{1.00} &  - \cellcolor[gray]{1.00} &  9 \cellcolor[gray]{0.90} &  2 \cellcolor[gray]{0.98} &  0 \cellcolor[gray]{1.00} &  0 \cellcolor[gray]{1.00} \\
LOC & 10 \cellcolor[gray]{0.90} & 10 \cellcolor[gray]{0.89} & 10 \cellcolor[gray]{0.90} &  7 \cellcolor[gray]{0.93} & 11 \cellcolor[gray]{0.88} & 16 \cellcolor[gray]{0.83} &  - \cellcolor[gray]{1.00} & 15 \cellcolor[gray]{0.84} &  0 \cellcolor[gray]{1.00} &  8 \cellcolor[gray]{0.92} \\
MNR &  0 \cellcolor[gray]{1.00} &  0 \cellcolor[gray]{1.00} & 22 \cellcolor[gray]{0.78} & 28 \cellcolor[gray]{0.71} & 26 \cellcolor[gray]{0.73} &  0 \cellcolor[gray]{1.00} &  4 \cellcolor[gray]{0.95} &  - \cellcolor[gray]{1.00} &  0 \cellcolor[gray]{1.00} & 33 \cellcolor[gray]{0.67} \\
PNC &  0 \cellcolor[gray]{1.00} &  0 \cellcolor[gray]{1.00} & 10 \cellcolor[gray]{0.90} &  7 \cellcolor[gray]{0.93} &  0 \cellcolor[gray]{1.00} &  0 \cellcolor[gray]{1.00} &  4 \cellcolor[gray]{0.95} &  2 \cellcolor[gray]{0.98} &  - \cellcolor[gray]{1.00} &  0 \cellcolor[gray]{1.00} \\
TMP &  3 \cellcolor[gray]{0.97} &  2 \cellcolor[gray]{0.97} &  2 \cellcolor[gray]{0.97} &  7 \cellcolor[gray]{0.93} & 38 \cellcolor[gray]{0.62} & 33 \cellcolor[gray]{0.67} & 23 \cellcolor[gray]{0.76} & 13 \cellcolor[gray]{0.87} &  0 \cellcolor[gray]{1.00} &  - \cellcolor[gray]{1.00} \\
\end{tabular}
}
\caption{Confusion matrix for labeling errors. Each cell shows the percentage of predicted labels for each gold label.}
\label{EXP:matrix}
\end{table}

\paragraph{Labeling Confusion}
Table \ref{EXP:matrix} shows a confusion matrix of our model for the most frequent labels. We only consider predicted arguments that match gold span boundaries. Compared with the previous work~\cite{he2017deep}, our model still confuses ARG2 with AM-DIR, AM-LOC and AM-MNR, but to a lesser extent. This indicates that our model has some advantages on such difficult adjunct distinction~\cite{kingsbury2002adding}.

\section{Related work}
\paragraph{SRL}
Gildea and Jurafsky~\shortcite{gildea2002automatic} developed the first automatic semantic role labeling system based on FrameNet. Since then the task has received a tremendous amount of attention. The focus of traditional approaches is devising appropriate feature templates to describe the latent structure of utterances. Pradhan et al.~\shortcite{Pradhan-Jurafsky-Conll2005}; Surdeanu et al.~\shortcite{Surdeanu-Aarseth-ACL2003}; Palmer, Gildea, and Xue~\shortcite{Palmer-Xue-2010} explored the syntactic features for capturing the overall sentence structure. Combination of different syntactic parsers was also proposed to avoid  prediction risk which was introduced by Surdeanu et al.~\shortcite{Surdeanu-Aarseth-ACL2003}; Koomen et al.~\shortcite{Koomen-Yih-CoNLL2005}; Pradhan et al.~\shortcite{Pradhan-CoNLL2013}.

Beyond these traditional methods above, Collobert et al.~\shortcite{Collobert-Ronan-JMLR2011} proposed a convolutional neural network for SRL to reduce the feature engineering. The pioneering work on building an end-to-end system was proposed by Zhou and Xu~\shortcite{zhou2015end}, who applied an 8 layered LSTM model which outperformed the previous state-of-the-art system. He et al.\shortcite{he2017deep} improved further with highway LSTMs and constrained decoding. They used simplified input and output layers compared with Zhou and Xu~\shortcite{zhou2015end}. Marcheggiani, Frolov,  Titov~\shortcite{marcheggiani2017simple} also proposed a bidirectional LSTM based model. Without using any syntactic information, their approach achieved the state-of-the-art result on the CoNLL-2009 dataset.

Our method differs from them significantly. We choose self-attention as the key component in our architecture instead of LSTMs. Like He et al.~\shortcite{he2017deep}, our system take the very original utterances and predicate masks as the inputs without context windows. At the inference stage, we apply argmax decoding approach on top of a simple logistic regression while Zhou and Xu~\shortcite{zhou2015end} chose a CRF approach and He et al.~\shortcite{he2017deep} chose constrained decoding. This approach is much simpler and faster than the previous approaches.

\paragraph{Self-Attention} Self-attention have been successfully used in several tasks. Cheng, Dong, and Lapata~\shortcite{cheng2016long} used LSTMs and self-attention to facilitate the task of machine reading. Parikh et al.~\shortcite{parikh2016decomposable} utilized self-attention to the task of natural language inference. Lin et al.~\shortcite{lin2017structured} proposed self-attentive sentence embedding and applied them to author profiling, sentiment analysis and textual entailment. Paulus, Xiong, and Socher~\shortcite{paulus2017deep} combined reinforcement learning and self-attention to capture the long distance dependencies nature of abstractive summarization. Vaswani et al.~\shortcite{vaswani2017attention} applied self-attention to neural machine translation and achieved the state-of-the-art results. Very recently, Shen et al.~\shortcite{shen2017disan} applied self-attention to language understanding task and achieved the state-of-the-art on various datasets. Our work follows this line to apply self-attention for learning long distance dependencies. Our experiments also show the effectiveness of self-attention mechanism on the sequence labeling task.

\section{Conclusion}
We proposed a deep attentional neural network for the task of semantic role labeling. We trained our SRL models with a depth of $10$ and evaluated them on the CoNLL-2005 shared task dataset and the CoNLL-2012 shared task dataset. Our experimental results indicate that our models substantially improve SRL performances, leading to the new state-of-the-art.

\section*{Acknowledgements}
This work was done while the first author's internship at Tencent Technology. This work is supported by the Natural Science Foundation of China (Grant No. 61573294, 61303082, 61672440), the Ph.D. Programs Foundation of Ministry of Education of China (Grant No. 20130121110040), the Foundation of the State Language Commission of China (Grant No. WT135-10) and the Natural Science Foundation of Fujian Province (Grant No. 2016J05161). We also thank the anonymous reviews for their valuable suggestions.

\bibliography{srl}
\bibliographystyle{aaai}

\end{document}